\documentclass[twoside,leqno,twocolumn]{article}
\usepackage[utf8]{inputenc}
\usepackage[letterpaper]{geometry}
\usepackage{ltexpprt}%
\usepackage{graphicx}
\usepackage[pdfborder={0 0 0}]{hyperref}
\usepackage{tikz}
\usepackage[textsize=tiny,disable]{todonotes} %

\usepackage{mathtools}
\usepackage{import}
\usepackage[noend]{algorithmic}
\usepackage{algorithm}
\usepackage{caption,subcaption}

\usepackage{amsmath,amssymb}

\usepackage{longtable} %
\usepackage{multirow}
\usepackage{lscape}

\usepackage[style=numeric,giveninits=true,backend=biber]{biblatex}
\AtEveryBibitem{\clearfield{pages}} %
\newcommand*\rotatelow{\rotatebox{50}} %
\usepackage[figuresleft]{rotating}
\newif\ifBlind\Blindtrue\Blindfalse  %

\newcommand{\reffig}[1]{Fig.~\ref{#1}}
\newcommand{\reftab}[1]{Table~\ref{#1}}

\newcommand{\refalg}[1]{Alg.~\ref{#1}}

\newcommand{\snorm}[1]{\left\lVert\smash{#1}\mathstrut\right\rVert}

\newcommand{\sgn}{\ensuremath{\operatorname{sign}}}

\DeclareMathOperator*{\argmin}{\arg\min} %
\DeclareMathOperator*{\argmax}{\arg\max}
\DeclareMathOperator*{\diag}{\operatorname{diag}}

\newcommand{\AUPRC}{\ensuremath{\operatorname{AUPRC}}}
\newcommand{\AUROC}{\ensuremath{\operatorname{AUROC}}}

\newcommand{\AveP}{\ensuremath{\operatorname{AveP}}}
\newcommand{\AdjAveP}{\ensuremath{\operatorname{Adj AveP}}}
\newcommand{\rank}{\ensuremath{\operatorname{rank}}}

\newcommand\relatedversion{}

\begin{document}
\title{\Large LOSDD: Leave-Out Support Vector Data Description for Outlier Detection\relatedversion}
\ifBlind\else
\author{
Daniel Boiar%
\and
Thomas Liebig%
\and
Erich Schubert%
\thanks{TU Dortmund University, Dortmund, Germany,
\nolinkurl{{daniel.boiar,erich.schubert,thomas.liebig}@tu-dortmund.de}}
}%
\fi
\date{}

\maketitle

\begin{abstract} \small\baselineskip=9pt

Support Vector Machines have been successfully used for one-class classification
(OCSVM, SVDD) when trained on clean data, but they work much worse on dirty data:
outliers present in the training data tend to become support vectors, and
are hence considered ``normal''.
In this article, we improve the effectiveness to detect outliers in dirty training
data with a leave-out strategy: by temporarily omitting one candidate at a time,
this point can be judged using the remaining data only.
We show that this is more effective at scoring the outlierness of points than
using the slack term of existing SVM-based approaches.
Identified outliers can then be removed from the data, such that outliers hidden
by other outliers can be identified, to reduce the problem of masking.
Naively, this approach would require training $N$ individual SVMs
(and training $\mathcal{O}(N^2)$ SVMs when iteratively removing the worst outliers one at a time),
which is prohibitively expensive. We will discuss that only support vectors need to be
considered in each step and that by reusing SVM parameters and weights, this incremental
retraining can be accelerated substantially.
By removing candidates in batches, we can further improve the processing time,
although it obviously remains more costly than training a single SVM. 

\end{abstract}

\section{Introduction}
Outlier detection is the task of detecting anomalous instances within a data set.
It is a key task in data mining, and belongs to
unsupervised learning, where no labels are available for the training data.
There are several subtle variations to this theme, which cause misunderstanding,
misuse of methods, and hence often poor performance.
While these could be differentiated with a slight change in terminology,
these differences are unfortunately not well established in the literature.
The term ``outlier detection'' is primarily used in the context where there
exist anomalies within the ``training'' data set, and the task is to score all objects
based on their anomalousness. Ideally, one would of course prefer a binary
classification (outlier or inlier), but this is all but reliable on typical
dirty data where such methods are used. In real use, the methods can
only generate candidates for further analysis -- whether or not
these instances are ``bad'' anomalies such as attacks,
or just ``unusual but normal'' objects cannot be distinguished in an
unsupervised setting. Nevertheless, inspecting these anomalous instances
can be helpful for (partially) labeling the data, to then later
use supervised classification instead.
Popular techniques for outlier detection include $k$-nearest neighbor
outlier detection~\cite{DBLP:conf/sigmod/RamaswamyRS00}
and the local outlier factor (LOF)~\cite{DBLP:conf/sigmod/BreunigKNS00},
but many more have been proposed (see, e.g., the survey \cite{DBLP:journals/sadm/ZimekSK12}).
Because the underlying problem is underspecificed, and there is little
\emph{meaningful} evaluation data available for the research community,
dozens of methods are proposed that do not offer reproducible benefits
\cite{DBLP:journals/datamine/CamposZSCMSAH16}
over standard methods such as KNN and LOF.

A different view arises in ``novelty detection'' or ``one class classification''.
Here, all training data is assumed to be normal, and new data points are
judged by their similarity to existing data. For example, we may have
sensor data from a machine that is working fine; we do not have training data
of possible future defects yet. %
This in turn bears similarities to change point detection and event detection
in data streams%
\ifBlind\else~\cite{DBLP:conf/icdm/SchubertWZ15}\fi.
Popular techniques include one-class support vector machines
(OCSVM)~\cite{DBLP:journals/neco/ScholkopfPSSW01}
and Support Vector Data Description (SVDD)~\cite{DBLP:journals/ml/TaxD04},
both of which are based on the support vector machine (SVM).
In this article, we will discuss the limitations of these approaches for
outlier detection in dirty data, and develop a method to make them perform
much better in this scenario.

\begin{figure}[bt]\centering
\includegraphics[width=.5\linewidth]{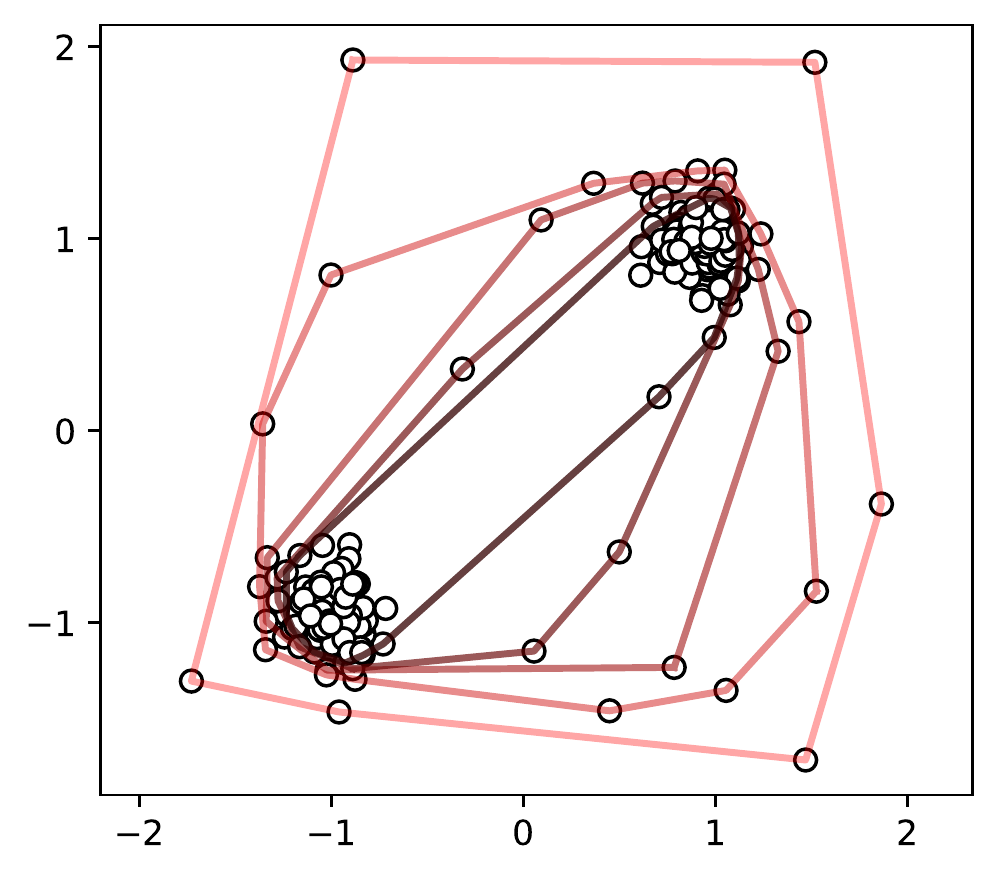}
\caption{Classic convex hull depth applied to two Gaussians and noise.}
\label{fig:convex-hull}
\end{figure}

A classic idea to describe data is the ``depth'' concept of J.~Tukey~\cite{Tukey:1975:MPD},
who used it as a visualization technique for bivariate data distribution.
This was connected to outlyingness by Barnett~\cite{doi:10.2307/2344839},
and later Donoho and Gasko~\cite{donoho1992},
and also forms the base for techniques such as the
Bagplot~\cite{doi:10.1080/00031305.1999.10474494}.
While ``convex hull peeling'' \cite{doi:10.2307/2344839} works well
as a visualization technique for data from a single multivariate,
but correlated distribution, it makes much less sense on data that
contains multiple clusters. In \reffig{fig:convex-hull} we show the first
five convex hulls of such a data set. We observe that this approach
misses anomalies in between of multiple clusters, and also lacks
discrimination among the members of the same hull, making this
classic statistical idea hard to use for anomaly detection
in multimodal data.

The approach that we propose in this paper is similar in spirit:
the data is circumscribed with layered hulls,
but our hulls can have an arbitrarily complex shape based on
support vectors in some kernel space. Because of slack, these hulls
are only approximate, and we can rank objects based on their position
with respect to the layered hulls.

\begin{figure}[tb]\centering
\begin{subfigure}{.48\linewidth}\centering
\includegraphics[width=\linewidth]{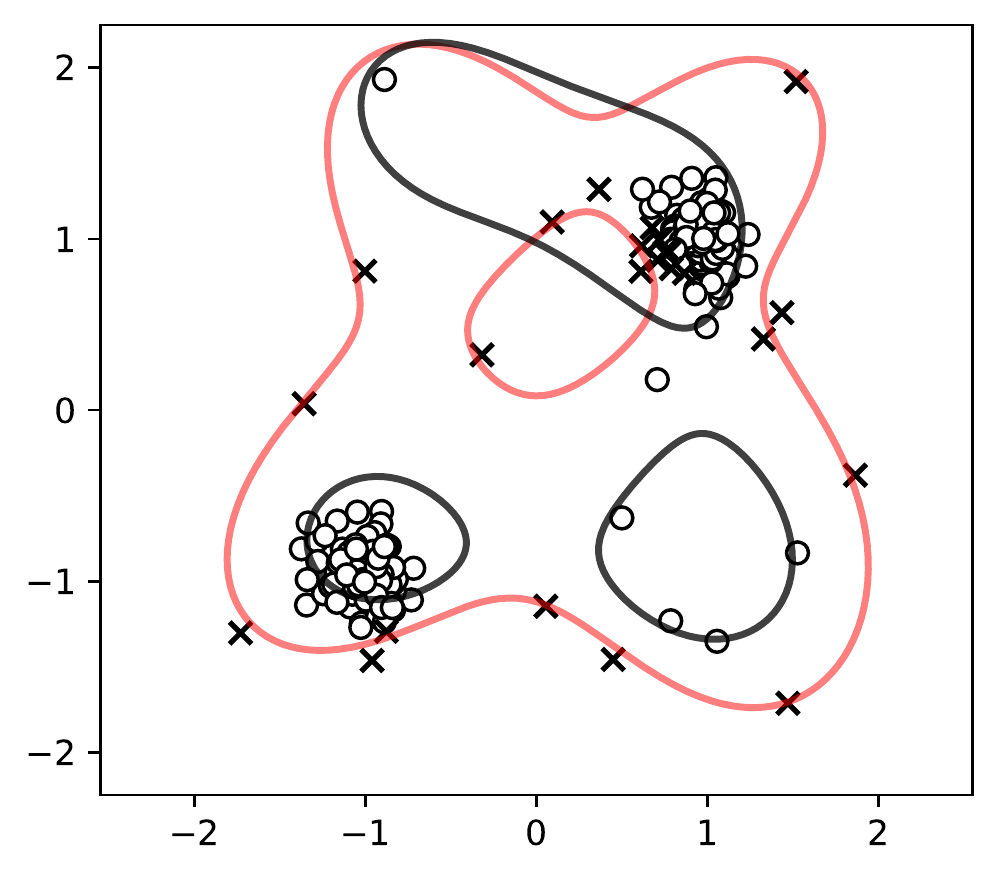}
\caption{Single OCSVM}
\label{fig:toy-intro-ocsvm}
\end{subfigure}%
\hspace*{\fill}%
\begin{subfigure}{.49\linewidth}\centering
\includegraphics[width=\linewidth]{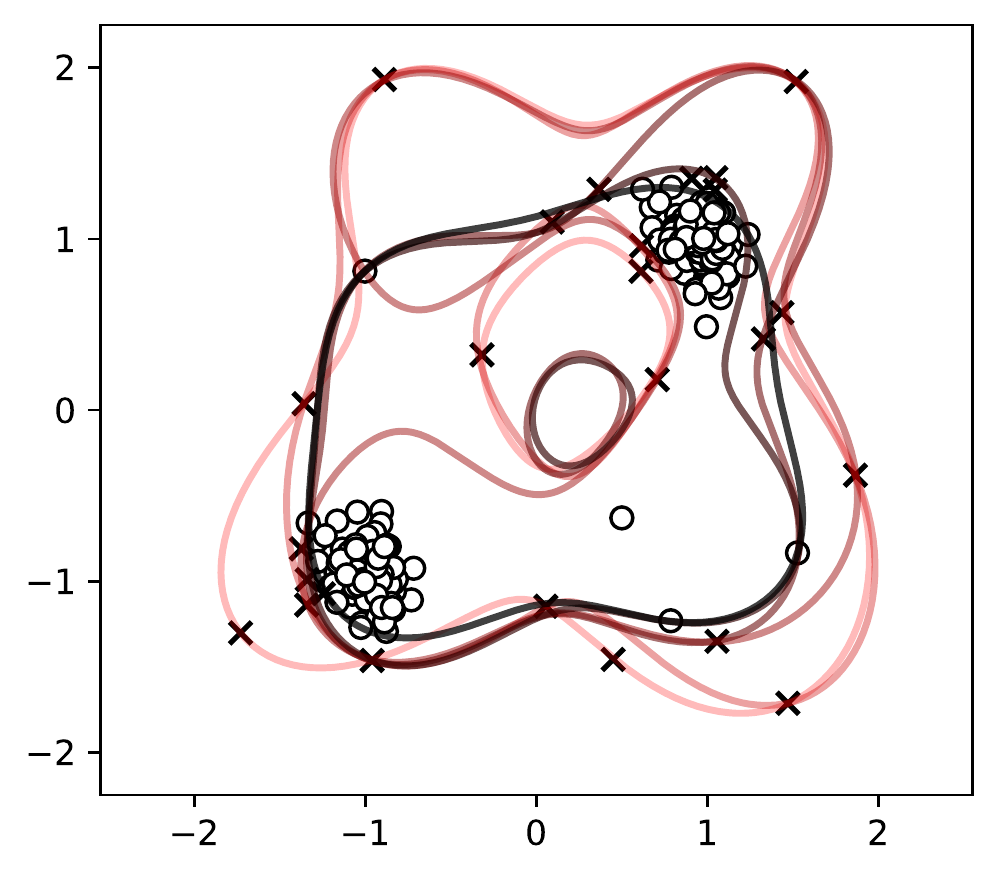}
\caption{Naive incremental}
\label{fig:toy-intro-naive}
\end{subfigure}
\\
\begin{subfigure}{.49\linewidth}\centering
\includegraphics[width=\linewidth]{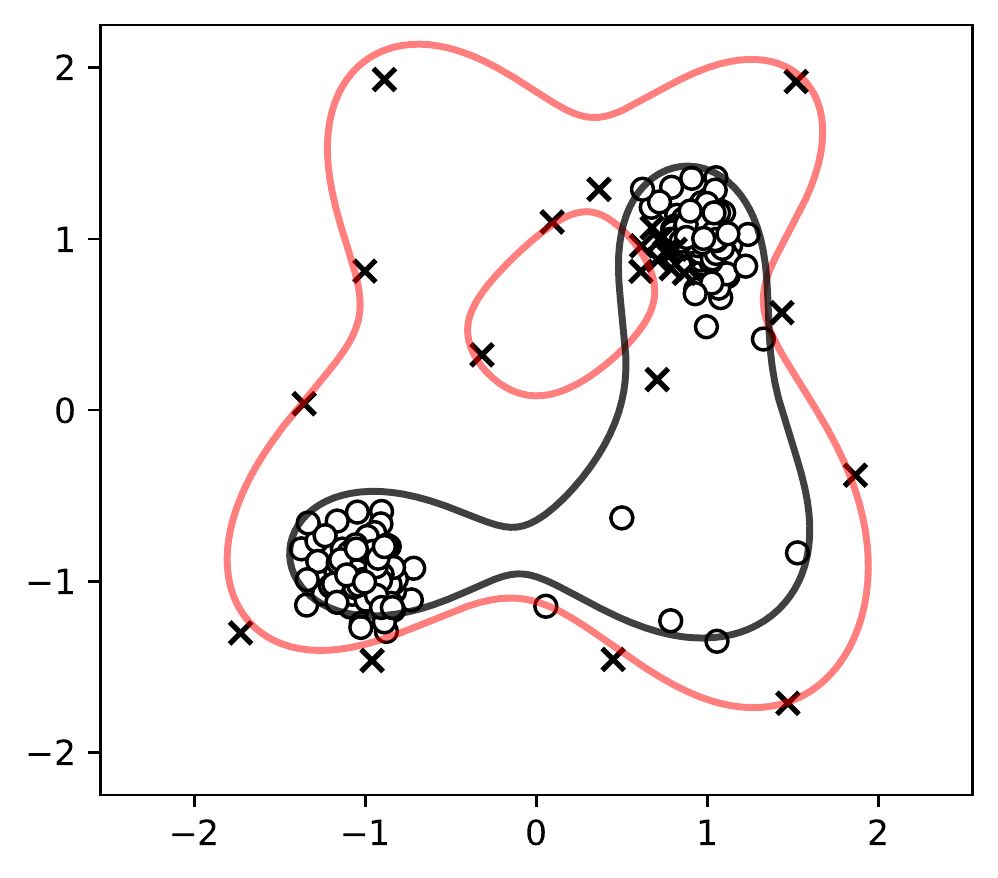}
\caption{Leave-one-out evaluation}
\label{fig:toy-intro-loo}
\end{subfigure}%
\hspace*{\fill}%
\begin{subfigure}{.49\linewidth}\centering
\includegraphics[width=\linewidth]{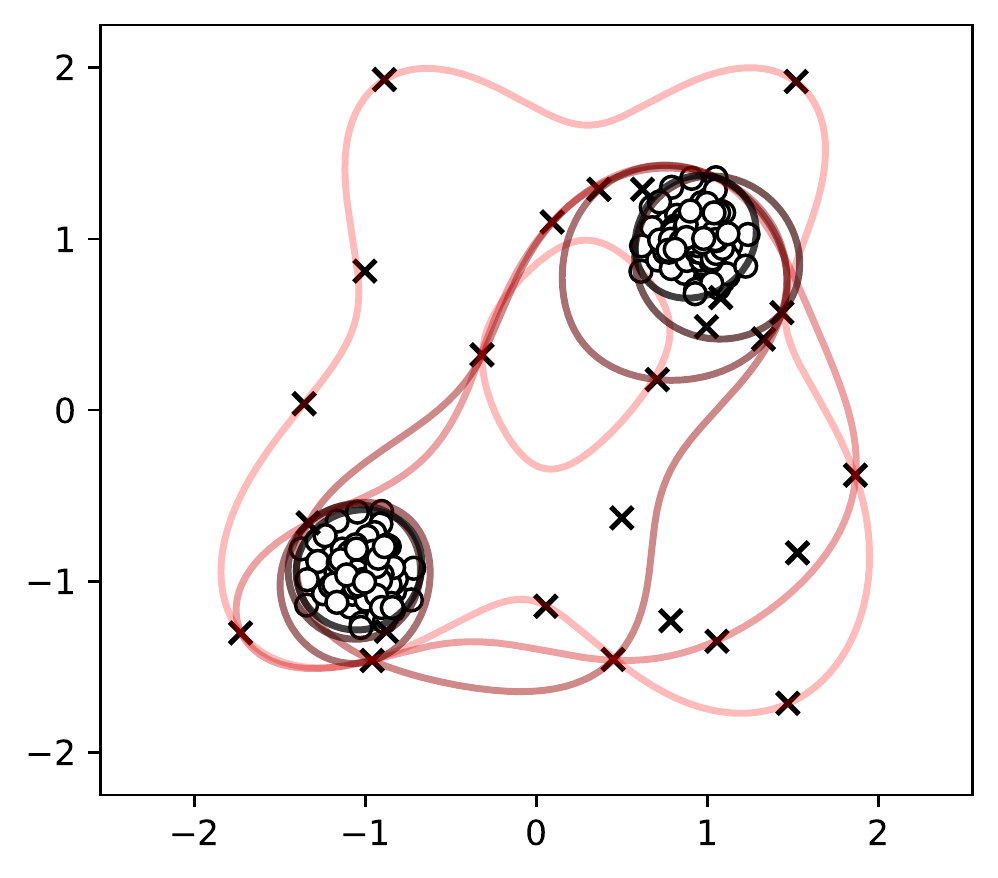}
\caption{LOSEVM}
\label{fig:toy-intro-losevm}
\end{subfigure}
\caption{Motivational example: removing outliers using support vector machines}
\label{fig:toy-intro}
\end{figure}

When using one-class support vector machines on ``dirty data''
(containing anomalies), we observed that these SVMs tend to overfit the anomalies in the data:
Even with slack, anomalous points as well as points in regions of low-density often become support vectors,
and hence may not be detected well. This is explainable by the fact that
one-class support vector machines assume that all data are inliers, and
attempt to detect novelties, not existing anomalies.

Removing the top 25 objects (by their distance to the separating hyperplane,
\reffig{fig:toy-intro-ocsvm}) does not yield
satisfactory results. While many of the outermost points were removed,
an entire group of low-density points in the bottom right (as well as
a quite isolated point in the top left remains). Instead, several points
close to the top-right cluster were removed. The red line in this figure
illustrates the initial decision boundary learned from the data, whereas
the black lines illustrate the decision boundary when trained on the
remaining data points only.

Based on the observation that removing outliers changes the decision
boundary, we came up with the idea of removing one outlier at a time,
then reevaluate. Repeating this procedure 25 times (hence again removing
25 points, but at the cost of training 25 support vector machines),
we obtain the results visualized in \reffig{fig:toy-intro-naive}.
We plot the decision boundaries after every 5 removals, shifting from
red to black during this procedure. The result has improved
in some places (the outlier in the top left is removed, and so is one
in the bottom right; fewer points concentrated at the top-right cluster
are removed), but still far from ideal: the final decision boundary
still surrounds much empty space, due to remaining anomalies in the
support vectors.
To improve the anomaly removal with SVMs, we observe that points to be scored
are better not part of the training data.
The result of training $N$ support vector machines (each one trained
on $N{-}1$ data points, in order to evaluate the left-out data point)
is shown in \reffig{fig:toy-intro-loo}. While the results have improved
over the classic OCSVM approach, the low-density points in the bottom
right remain undetected: they are still ``masked''
by their neighbor outliers.

Combining these two techniques finally yields much better results:
In \reffig{fig:toy-intro-losevm} we show the decision boundaries
after every 5 removals (beginning with the same initial decision
surface as all other plots). After removing 15 objects,
the two clusters have separated, and the outlier detection
result is much better. The explanation is that by removing one outlier
at a time, the next outliers in the group become exposed, and
we are able to remove such grouped outliers now.

Unfortunately, the run time of this approach now has become infeasible
for large data sets. A straightforward implementation %
has to train $N$ support vector machines at each iteration. If we need
to remove 1\% of the data objects, we hence have to train $\mathcal{O}(N^2)$
support vector machines, each of which has an empirical training cost
of $N^2$ to $N^3$. In this article, we discuss strategies
to improve this procedure, to reduce run time to an acceptable
level for many data sets.

The fully unsupervised dirty outlier detection task which is addressed here is formalized by this problem: We are given $N$ points with no label information. A small unknown proportion of this set are outliers, such that no semi-supervised training could be used for tuning the hyperparameters. The type of outlier is not specified further than that an outlier is an object that is inconsistent with the normal points \cite{barnett1984outliers}.

\section{Related Work}

Amer et al.~\cite{ACM:Amer13} observe that the outliers contribute most to the decision boundary and
modify the objective function of the OCSVM, scaling the distances to the center of gravity
with the Lagrange multipliers $\alpha$, such that outliers with
a high distance to the center of gravity get a reduced $\alpha$.
The authors claim to obtain a more balanced solution this way, computed to the solution of the original optimization problem.
However, this approach ``breaks'' the original SVM intuition of minimizing slack,
effectively allowing objects to have much more slack.
Our approach attempts to follow closely the SVM optimization problem,
and instead use the assumption that some outlier objects need to be removed from the ``dirty'' data set entirely.

Weston~\cite{DBLP:conf/ijcai/Weston99} proposed a variant of SVMs that minimizes the leave-one-out
classification error.

Joachims~\cite{DBLP:conf/icml/Joachims00} proposes a computationally efficient $\xi\alpha$-estimator which predicts
the performance of a binary classification SVM with the leave-one-out evaluation. %
This estimator does not score individual points, but estimates the aggregated error from the  
slack $\xi$ and the Lagrange multipliers $\alpha$ of the converted SVM.
Joachims also suggests to additionally evaluate certain points in the leave-one-out procedure to increase the accuracy of the estimate.
In the proofs, steps of retraining of a SVM are used. 
We use these ideas for retraining a SVM in the one-class scenario.

Tax and Duin~\cite{DBLP:journals/ml/TaxD04} refer to the leave-one-out procedure of the SVDD to estimate the error and thus gives instructions to choose the parameter for $\gamma$ and $C$.
Among other things, the parameterization is further discussed by Xiao et al.~\cite{DBLP:journals/tcyb/XiaoWX15}.

Ruff et al.{}~\cite{DBLP:conf/icml/RuffGDSVBMK18} proposed Deep One-Class Classification (DeepSVDD),
but deep learning techniques usually require massive hyperparameter tuning, and the method performed significantly worse
than all other unsupervised methods in the recent ADBench benchmark~\cite{han2022adbench}:
``We also note that some DL-based unsupervised methods like DeepSVDD and DAGMM
are surprisingly worse than shallow methods. Without the guidance of label information, DL-based
unsupervised algorithms are harder to train (due to more hyperparameters) and more difficult to tune
hyperparameters, leading to unsatisfactory performance.''
Deep learning methods such as DeepSVDD and REPEN~\cite{DBLP:conf/kdd/PangCCL18} appear to
require semi-supervision and particular data types such as images to work well.

\begin{figure}[tb]
\begin{subfigure}{.33\linewidth}\centering
\includegraphics[width=\linewidth]{toyexample/toy-loo-25-1-1.pdf}
\caption{Top 25 at once}
\label{fig:toy}
\end{subfigure}%
\hfill%
\begin{subfigure}{.33\linewidth}\centering
\includegraphics[width=\linewidth]{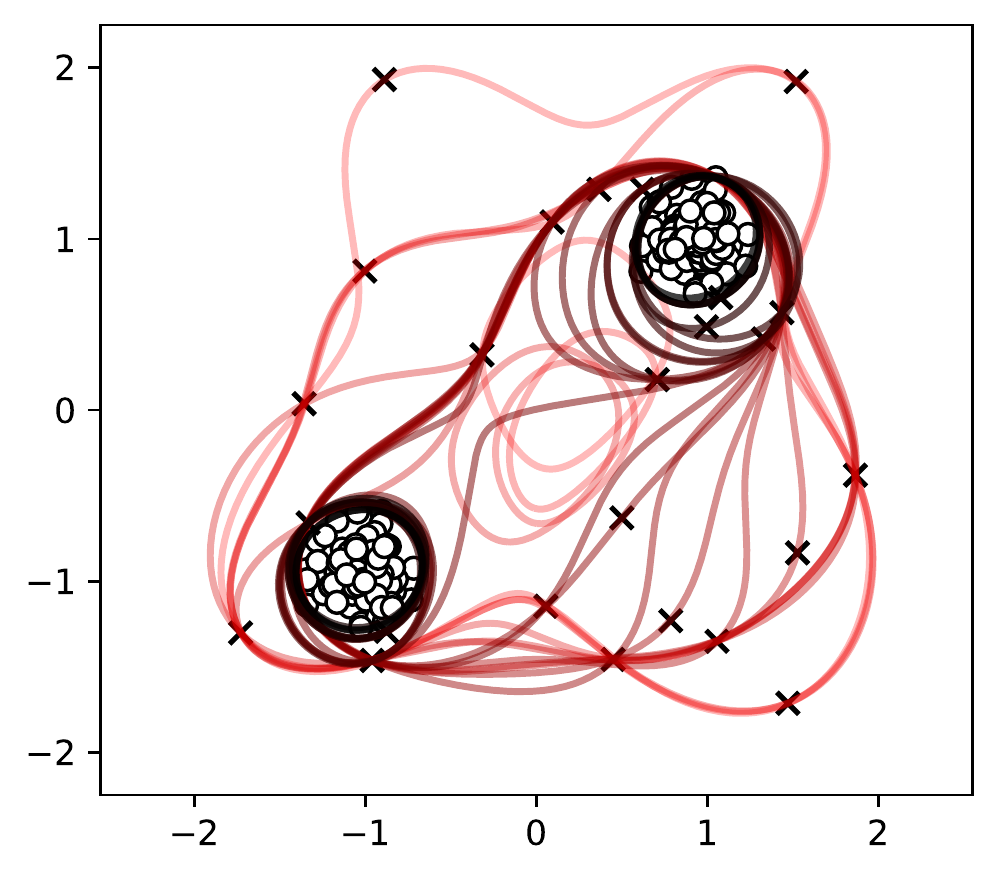}
\caption{One at a time}
\label{fig:toy}
\end{subfigure}%
\hfill%
\begin{subfigure}{.33\linewidth}\centering
\includegraphics[width=\linewidth]{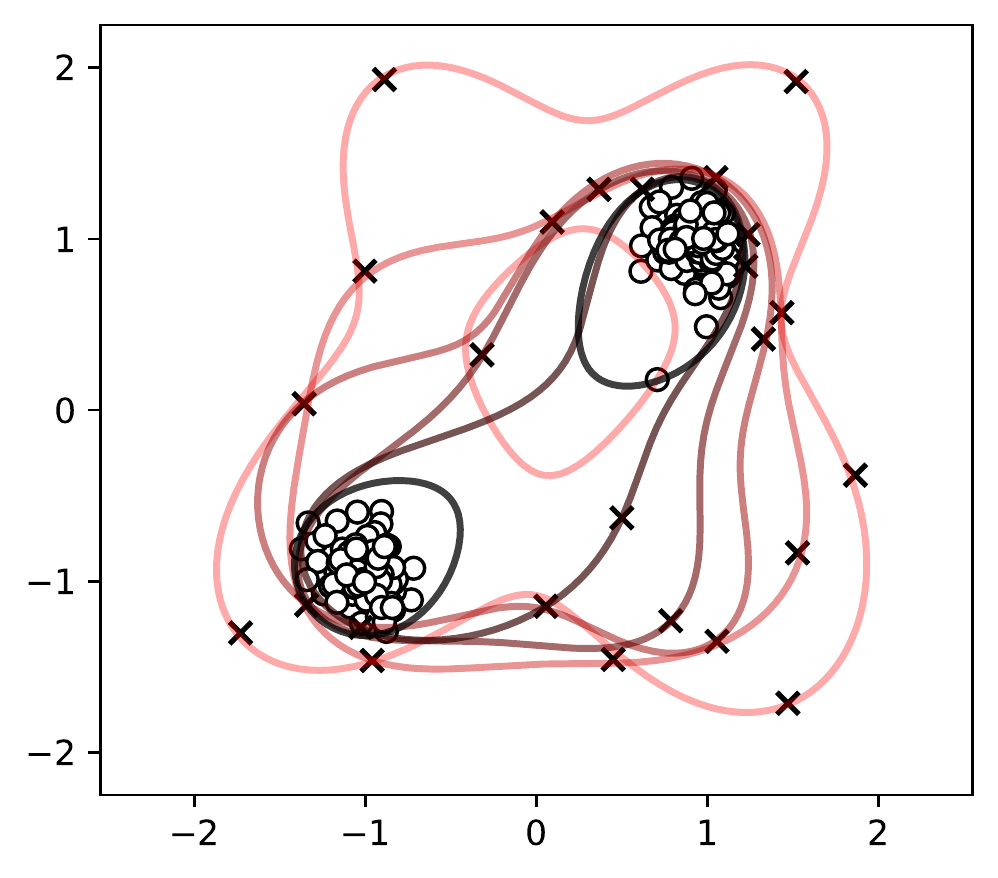}
\caption{Batches of 5}
\label{fig:toy}
\end{subfigure}
\caption{Removing outliers with leave-one-out scoring}
\label{fig:toy-loo}
\end{figure}

\section{Foundations}
In this section we describe how the OCSVM and the SVDD learn, decide, achieve their global optimality and train efficiently.
Afterward, the new algorithm is presented, including a batchwise strategy that is motivated by \reffig{fig:toy-loo}.

\subsection{One-Class Support Vector Machines} %
The idea of the OCSVM \cite{DBLP:journals/neco/ScholkopfPSSW01} is to find a hyperplane
$H{:{}}w\cdot \Phi(x){-}b{=}0$, $w{\in}F$, $b{\in}R$ that maximally separates the data $X$
from the origin in a feature space $F$ obtained by projection $\Phi\colon X\rightarrow F$. %
The weights $w$ and the bias $b$ can be obtained by solving the quadratic optimization problem:
\begin{align}
    \min_{w{\in} F, \xi{\in}\mathbb{R}^N, b{\in}\mathbb{R}} & \tfrac{1}{2}\snorm{w}^2+\tfrac{1}{\nu}\tfrac{1}{N}\sum\nolimits_{i=1}^N\xi_i-b
\label{OCSVMoptimierungsproblem}\\ %
     \text{subject to\enskip}& w\cdot \Phi(x_i)\geq b-\xi_i \ \forall i\in\{1,\dots,N\}, \notag\\
     & \xi_i\geq 0 \ \forall i\in\{1,\dots,N\} \;.\notag
\end{align}
As usual with SVMs, $\Phi$ is only this is only used implicitly using a scalar product kernel $K(x_i,x_j)=\Phi(x_i)^T \Phi(x_j)$.
The slack variables $\xi_i$ allow tolerating an error in the optimization to favor a simper solution
rather than overfitting the data, while the parameter $\nu{\in}(0, 1)$ controls the amount of slack.
To use a uniform notation with SVDD below, we define $C{:=}\frac{1}{\nu}\frac{1}{N}$.
If a test sample is located outside the smaller separated space, it is classified as an outlier, otherwise as inlier.
The function $f(x){=}\frac{w\cdot\Phi(x)-b}{\snorm{w}}$ determines the shortest orientated distance from the hyperplane~$H$ to the point~$x$.
The sign of $f$ indicates the side on which $x$ is located.
To solve \eqref{OCSVMoptimierungsproblem}, the Lagrange method with multipliers $\alpha_i,\beta_i\geq 0$ %
is used to obtain
\begin{align}
L(w,\xi,b,\alpha,\beta)=\tfrac12\snorm{w}^2{+}C\sum\nolimits_{i=1}^N\xi_i{-}b
\label{Lagrange}\\
{-}\sum\nolimits_{i=1}^N\alpha_i\big(w{\cdot}\Phi(x_i){-}b{+}\xi_i\big){-}\sum\nolimits_{i=1}^N \beta_i\xi_i
\;.
\notag
\end{align}
Zeroing the derivatives of $L$ regarding all variables yields the conditions
\begin{align}
w {=}\sum_{i=1}^N\alpha_i\Phi(x_i)
\;, \enskip%
\alpha_i {=} C{-}\beta_i\leq C
\;, \enskip%
1=\sum_{i=1}^N \alpha_i
\;. \label{loesungVonLagrange}
\end{align}
The observations $x_i$, for which $\alpha_i{>}0$, are called support vectors (SV).
Substituting these conditions in the Lagrange function $L$ and using the Wolfe Dual Theorem~\cite{10.5555/39857}
leads to the dual optimization problem:
\begin{align}
\min\limits_{\alpha} \; & \frac{1}{2}\sum\nolimits_{i=1}^N\sum\nolimits_{j=1}^N \alpha_i\alpha_j K(x_i,x_j)
\label{OCSVMoptimierungsproblemDual}
\\
\text{subject to} \; & 0\leq \alpha_i\leq C \text{ and } \sum\nolimits_{i=1}^N \alpha_i=1
\notag\\
\shortintertext{ with the dual decision function }
  f(x)&=\sgn\big(\sum\nolimits_{i=1}^N\alpha_i K(x_i,x)-b\big) \;.
\label{AbstandOCSVMPred}
\\
\shortintertext{ The bias $b$ is calculated by placing a SV $x_{SV}$ with $\alpha_{SV}{<}C$ in the hyperplane $H$: }
b=&
w\cdot\Phi(x_{SV})=\sum\nolimits_{j=1}^N\alpha_j K(x_j,x_{SV}) \;.
\notag
\end{align}

The kernel function $K$ was introduced to indirectly calculate the projection using a scalar product. 
A popular kernel is the radial basis function~(RBF) kernel, $K(x,y) = e^{-\gamma\snorm{x-y}^2}$.
The parameter $\gamma$ controls the influence similar to the bandwidth in kernel density estimation.
A high $\gamma$ leads to a small bandwidth and thus to an over-adjustment of the decision boundary at each point.

\subsection{Support Vector Data Description (SVDD)}
A similar approach to the OCSVM is to describe the data by a ball with minimal volume \cite{DBLP:journals/ml/TaxD04}. 
A point inside the ball belongs to the inliers and a point outside the ball is defined as an outlier. 
The ball is defined by its center~$a$ and radius~$R$, where the radius is minimized under the
constraint that the distance to the center for all data points is bounded by the radius. 
Again, slack variables $\xi_i$ are added to allow some error, controllable by the meta parameter $C$. 
\begin{align}
  \min\limits_{R\in \mathbb{R},a\in\mathbb{R}^d}\enskip& \frac{1}{2}R^2+C\sum\nolimits_{i=1}^N\xi_i
\label{SVDDoptimierungsproblem}\\
    \text{subject to}& \snorm{x_i-a}^2\leq R^2 +\xi_i \ \forall i\in\{1,\dots,N\}\;, \notag\\
    & \xi_i\geq 0 \ \forall i\in\{1,\dots,N\} \;. \notag
\end{align}
This is solved using the Lagrange method, where $\alpha_i{\geq} 0$ are introduced for radius violations
and $\gamma_i{\geq} 0$ for slack to get the Lagrange function
\begin{align*}
L(R,a,\alpha,\gamma,\xi) = R^2+C\sum\nolimits_{i=1}^N \xi_i \\- \sum\nolimits_{i=1}^N \alpha_i [R^2+\xi_i -(\snorm{x_i-a}^2)]-\sum\nolimits_{i=1}^N \gamma_i\xi_i
\;.
\end{align*}
Zeroing the derivatives regarding all variables results~in
\begin{align}
\sum\limits_{i=1}^N \alpha_i {=} 1
, %
a {=}\sum\limits_{i=1}^N \alpha_i x_i
, %
0 {\leq} \alpha_i {\leq} C \ \forall i{\in}\{1,\dots,N\} \;.
\label{defCenter}
\end{align}
By introducing a projection into a feature space with the projection $\Phi$
and kernel function $K(x_i,x_j){=}\Phi(x_i)^T\Phi(x_j)$, we arrive at the dual optimization problem
\begin{align}
    \min\limits_{\alpha}\text{\hspace{10px}}& \sum\nolimits_{i=1}^N\sum\nolimits_{j=1}^N \alpha_i\alpha_jK(x_i,x_j) \notag
    -\sum\nolimits_{i=1}^N\alpha_iK(x_i,x_i)
\notag\\
      \text{s.t. \hspace{10px}}& 0\leq \alpha_i\leq C \ \forall i\in\{1,\dots,N\}, %
      \sum\nolimits_{i=1}^l \alpha_i = 1 \;. \notag
\end{align}
The result only depends on non-negative $\alpha_i{>}0$, and the 
corresponding points $x_i$ are the support vectors. 
Because the support vectors define the ball that sketches the data, the approach is therefore called Support Vector Data Description.
A point $z$ is an inlier iff 
\begin{align*}
f(z):=\snorm{\Phi(z)-\Phi(a)}^2 - R^2\leq 0.
\end{align*} 
This is equivalent to
\todo{Diesen Teil eindampfen auf die letzte Formel? vielleicht nicht mehr jetzt so kurz vor der deadline}%
\begin{align*}
K(z,z) - 2\sum\nolimits_{i=1}^N \alpha_i K(z,x_i) \notag\\\leq K(x_k,x_k) - 2 \sum\nolimits_{i=1}^N \alpha_iK(x_k,x_i)\;.
\end{align*}

\section{Leave-Out Support Vector Machines}
When studying the decision boundary of a OCSVM or a SVDD, we observe that outliers are commonly part of the support vectors.
As previously observed by, e.g., Fletcher~\cite{10.5555/39857} and Amer et al.~\cite{ACM:Amer13}, outliers often have a large $\alpha_i$,
and hence a large influence on the decision function.
Both OCSVM and SVDD rely on the distance of a point to the decision boundary to identify outliers,
but exactly this value is influenced substantially by outliers.
Similar to nearest-neighbor based outlier detection, where we have to leave out the test point to calculate a useful outlier score, we would like to leave out the test point for the SVM-based methods.
The basic LO-SVM score of a point is the distance to a SVM trained on the remainder of the data set.
However, naively leaving out a SV (e.g.,~by temporarily setting $\alpha_i{=}0$ when testing $x_i$) changes the decision boundary in a non-negligible way,
as the boundary was optimized to satisfy $\sum_{i=1}^l \alpha_i = 1$. Simply removing a point hence breaks the optimality criteria of the SVM.
Instead, we need to train a new SVM on all data except the test point to obtain a reliable boundary.
The converse is more helpful: leaving out a point that is not a support vector ($\alpha_i{=}0$) does not change the boundary,
we omit the proof for brevity.
\begin{theorem}\label{theorem:inlierDoesntChangeBorder}
 Let $\alpha_1,\dots,\alpha_N$ be the solution of an SVM where all points have been trained.
 Let $x_t$ be a point with $\alpha_t=0$. Then after leaving out the point $x_t$ the
 solution $\alpha_1,\dots,\alpha_{t-1},\alpha_{t+1},\dots,\alpha_N$ is optimal. 
\end{theorem}

To carry out this procedure for calculating the outlier score naively,
one would need to train $N$ SVMs with $N{-}1$ points in the training set each.
Instead, we propose the following improved approach: we first obtain an initial model using the entire data set.
Points with $\alpha_i{=}0$ can simply be scored using this initial model, and will by definition be inliers.
For the support vectors, we follow ideas from Joachims~\cite{DBLP:conf/icml/Joachims00}, and reuse this initial solution
to obtain better starting conditions for the optimizer.
When one SV is left out, its $\alpha{>}0$ needs to be distributed to other points to restore the condition $\sum_i \alpha_i = 1$
and obtain a feasible model on the remaining points.
This model is then optimized using a standard SVM solver using the gradient of the objective function.
Instead of recomputing all gradients, we can also try to adapt the gradients to the change in the $\alpha$,
which is easiest if the entire $\alpha$ is assigned to exactly one non-support vector.
A non-support vector is guaranteed to be able to take the entire $\alpha$, whereas assigning the weight
to an existing SV may violate the upper bound given by the regularization parameter $C$.
Even though this non-support vector may be a suboptimal choice, the common approach of finding the
most violating pair will usually transfer this weight to the appropriate points instead.
In standard SVM optimization, the gradient is obtained from the objective function
\begin{align*}
F(\alpha) =& \sum\limits_{i=1}^N\sum\limits_{j=1}^N \alpha_i\alpha_jK(x_i,x_j)-\sum\limits_{i=1}^N\alpha_iK(x_i,x_i) \\=& \alpha^T K \alpha - \diag(K)^T \alpha
\end{align*}
as $G:= \nabla F(\alpha) = K \alpha-\diag(K)$,
where $K$ is the matrix $(K(x_i,x_j))_{i,j\in\{1,\dots,N\}}$ and $\diag(K)$ its diagonal.
Thereby $\nabla F(\alpha)_i$ denotes the $i$th component of the gradient.
The gradient is used to find the pair $\alpha_i$, $\alpha_k$ that most violates the necessary conditions of being an optimum (KKT-conditions)
by checking $\min\nolimits_{\alpha_i<C}\nabla F(\alpha)_i\geq \max\nolimits_{\alpha_i> 0}\nabla F(\alpha)_i$.
After finding the most violating pair
\begin{equation}\label{mostViolatingPair}
\alpha_i=\argmax\limits_{\alpha_i>0}\nabla F(\alpha)_i, \quad \alpha_k=\argmin\limits_{\alpha_i<C}\nabla F(\alpha)_i,
\end{equation}
the previously calculated gradient is only partially recalculated (updated) by 
\begin{equation*}
G_j \leftarrow G_j + K_{ij}(\alpha_i^{\textit{new}}-\alpha_i^{\textit{old}}) + K_{kj}(\alpha_k^{\textit{new}}-\alpha_k^{\textit{old}}) \ \forall j. %
\end{equation*}
where $\alpha_i^{\textit{old}}$ denotes the value $\alpha_i$ found in \eqref{mostViolatingPair}, and $\alpha_i^{\textit{new}}$ denotes a better value for $\alpha_i$ to reduce violation of the necessary conditions. 
The latter can be calculated using, e.g., the Sequential Minimal Optimization algorithm \cite{DBLP:conf/nips/Platt98}.
When we distribute the Lagrange multiplier $\alpha_t$ of the left out point to a non-SV, we will have 
$\alpha_i^{\textit{new}} = \alpha_t$, $\alpha_i^{\textit{old}} = 0, \alpha_k^{\textit{new}}= 0, \alpha_k^{\textit{old}} = \alpha_t$. 
Thus the awakened point is assigned $\alpha_t$ and the remains of the left out point $x_t$ are cleared away.

For our implementation, we extend the widely used LIBSVM~\cite{DBLP:journals/tist/ChangL11}.
This involves paying attention to a number of technical implementation details,
such as resource management. We want to avoid recomputations among multiple runs of the solver
as good as possible, and this involves keeping many data structures to store the
kernel matrix cache, the gradients, and the $\alpha$ values in sync.
In particular, retaining the kernel matrix cache is beneficial for run time here,
but due to the design of LIBSVM (which constantly permutes the data set to have the support vectors
form a hot set) needs some attention.
To facilitate the removal of a point, we swap it to the end of the data set,
then temporarily reduce the data set size.
Instead of the original LIBSVM C code (or its one-to-one Java port),
we extend the more object-oriented adaptation in the ELKI framework~\cite{DBLP:conf/sisap/Schubert22},
in which we will make our code available. %

The LO-SVM technique can be combined with both OCSVM and SVDD, which we denote as
LOSOC and LOSDD below.

\section{Repeated and Batchwise Removal}
Above procedure of evaluating each point compared to a SVM trained on the remainder
yields the result seen in \reffig{fig:toy-intro-loo}, which still contained a number
of points masked (and some parts of the cluster in the top right removed instead).
Removing one support vector may expose additional outliers previously masked,
which eventually leads to the better result in \reffig{fig:toy-intro-losevm}.
Therefore, we propose to remove the most outlying support vector, and then repeat the procedure.
It makes little sense to repeat this until all points have been removed,
but the user may choose to remove a fixed number of outliers or a fixed share of the data set (e.g., 1\%).
However, for each such iteration, we again have to train $O(N)$ additional support vector machines,
making this fairly expensive (even when reusing the kernel cache and the internal data structures of the optimizer as before).
To lessen the computational cost, a simple idea is to remove a batch of objects instead of a single object
in each step. For small data sets, we may remove one object at a time, but for larger data sets we might, e.g.,
remove the 10 most outlying points (with the highest distance according to the approach above) at once.
The amount of batches that are to be removed, is a user-parameter called $b$.
The parameter for the number of points that are removed in total is called $R$ and is set to $b$ by default. With this specification each batch removes $\frac{R}{b}$ points per batch.
In the following the pseudo code \refalg{alg:LOSEVM} is shown. 
In each batch round the method iterates over each SV, swap it out, updates $\alpha, G$ and call the solver for the final iteration correcting the redistributed $\alpha$ and $G$. 
\begin{algorithm}[H]
\caption{LO-SVM(R: remove total, b: batchsize)}
\label{alg:LOSEVM}
\begin{algorithmic}[1]
		\STATE{init $\alpha,K$; svm.train($\alpha,K$); $H\leftarrow heap(size=\frac{R}{b})$ }
        \STATE{score$(x_i) \leftarrow $model.predict$(x_i)\ \forall i\in\{1,\dots,l\}$}
		\FOR{$batch \in\{1,\dots,b\}$}        
        	\FOR{$x_t$ is $SV$ in model}
                	\STATE $l\leftarrow N-1-$removed; swap($SV,l$);
                	\STATE $\alpha^{old}_t \leftarrow \alpha_t$;$\alpha_t \leftarrow 0$;updateG;$\alpha \leftarrow$ distr($\alpha^{old}_t$, $\alpha$)
                    \STATE solver.optimize($\alpha, K, G)$
                    \STATE score$(x_{t})\leftarrow $model.predict$(x_t)$;$H$.add(score$(x_t))$
            \ENDFOR
            \WHILE{$H\neq \emptyset$} %
            	\STATE $x_t\leftarrow H$.pop()
            	\STATE removed$++$;$l\leftarrow N-1-$removed; swap($SV,l$)
                \STATE $\alpha^{old}_t \leftarrow \alpha_t$;$\alpha_t \leftarrow 0$;updateG;$\alpha \leftarrow$ distr($\alpha^{old}_t$, $\alpha$)
            \ENDWHILE
        \STATE solver.optimize($\alpha, K, G)$
        \ENDFOR
\end{algorithmic}
\end{algorithm}

\section{Kernel Parameterization}
Although great flexibility is offered by the hyperparameters $C$/$\nu$ and $\gamma$~(from the RBF kernel),
it is not possible to choose optimal parameters in the unsupervised scenario where we lack a validation possibility
(in classification, these parameters are typically set by doing a grid search and cross-validation).
Hence we suggest to choose a small value for $\nu$ (e.g., $\nu{=}\tfrac{1}{n}$ corresponding to $C{=}1$):
this will cause the SVM to closely fit the data, meaning that it is more likely to overfit the boundary to single points,
which however does not harm our approach as much because of the leave-out approach where the test sample is not in the training set anymore.

The kernel parameter $\gamma$ also has an important influence; some heuristics are discussed by Xiao et al.~\cite{DBLP:journals/tcyb/XiaoWX15}. 
For example, Evangelista et al.'s~\cite{DBLP:conf/icann/EvangelistaES07} variation coefficient takes the variance of the kernel matrix into account and uses gradient ascent with multiple calculations of the kernel matrix.
Other examples are geometric approaches, sometimes in combination with the k-nearest neighbor learner.
We are particularly interested in the Gaussian RBF kernel, which uses a Gaussian distribution.
The close relation of one-class support vector machines with the RBF kernel to the kernel density estimation problem has been discussed by Mu\~{n}oz and Mogueraza~\cite{DBLP:conf/ciarp/MunozM04}.
For kernel density estimation (KDE),
there are many publications (e.g.,~\cite{scott15,silverman86,gramacki18}) that describe how to choose the bandwidth parameter, 
which lead to three main categories \cite{gramacki18} of bandwidth selectors:
(1)~normal scale or rule of thumb selectors, (2)~plug-in selectors, and (3)~cross validation selectors. %
Normal scale methods use the variance in the data, %
plug-in selectors minimize a loss function (e.g., the mean integrated square error), and often use a pilot function and a pilot kernel to include assumptions, which we do not have here. 
The cross validation selectors estimate their kernel density function on a subset of the data and test on another subset (e.g., leave-one-out) while optimizing the kernel bandwidth parameter over a loss function between the kernel density function on the training subset and its true density distribution, which is deducted. 
The latter would need multiple complete RBF kernel matrix calculations to find a good parameter, does not provide a closed-form equation, and can get stuck in locally optimal solutions.
For simplicity, we compare three very popular and easy-to-use methods, which belong to the normal scale selectors:
(1)~Scott's rule of thumb \cite{scott15}: $\sigma {=} n^{-\frac{1}{d+1}}\cdot \textit{std}$,
(2)~sklearn heuristic \cite{scikit-learn}: $\sigma {=} 2\cdot d^{\frac{1}{2}}\cdot \textit{std}$, and
(3)~Silverman's rule of thumb \cite{silverman86}:  $\sigma {=} (n\cdot (d + 2) \cdot \tfrac14)^{-\frac{1}{d+4}}\cdot \textit{std}$,
all of which rely on the data dimensionality $d$, the number of points $n$, and the data variance.
The formulas are converted from the bandwidth $\sigma$ notation from the literature to the $\gamma$ notation with the formula $\gamma{=}\tfrac{1}{2\sigma^2}$:
(1)~Scott: $\gamma {=} \tfrac12 n^{\frac{2}{d+4}} \cdot \frac{1}{\textit{var}}$,
(2)~sklearn: $\gamma {=} \frac{1}{d} \cdot \frac{1}{\textit{var}}$, and
(3)~Silverman: $\gamma {=} \tfrac12 (n \cdot (d + 2) \cdot \tfrac14)^{\frac{2}{d + 4}} \cdot \frac{1}{\textit{var}}$.

\section{Experiments}
Campos et al.~\cite{DBLP:journals/datamine/CamposZSCMSAH16} studied the performance of nearest-neighbor
based outlier detectors on a selection of data sets commonly used in literature. In contrast to later
outlier detection data collection efforts, the authors emphasize that not all data sets from,
e.g., the UCI repository \cite{Dua2019UCI} contain meaningful outliers; and even data sets commonly
used in literature may be a poor choice for outlier detection evaluation. For downsampled data
sets (unfortunately a commonplace hack to obtain ``outliers'' from classification data),
Campos et al.~provide reproducible folds for download which already have been
normalized, cleaned from duplicates and missing values. Additionally, we standardized the data. Nevertheless, not all of the data sets
have proven to yield useful results with all algorithms: some are too easy, but many too difficult
to solve without labels and without supervised hyperparameter tuning.
As seen in ADBench~\cite{han2022adbench}, blindly using as many data sets as possible will often just lead to
the outcome that no method is ``none of the unsupervised methods is
statistically better than the others'' to all the others.
Here, we focus on data sets that have shown to be feasible, yet difficult,
by Campos et al.~\cite{DBLP:journals/datamine/CamposZSCMSAH16} and
where we may be able to measure differences between the methods.
Nevertheless, given the poor availability of data sets actually labeled for outlier detection,
we cannot expect to see huge differences -- the field is still missing its ``imagenet moment''.

For evaluation, we use the popular measures Adjusted Average Precision and \AUROC{}.
Average Precision (\AveP) is the average precision at each outlier position in the ranking,
$\frac1{|O|}\sum_{o\in O} P@\rank(o)$ where $P@k = \frac{|\{o\in O\mid \rank(o)\leq k\}|}{k}$
is the precision at rank $k$.
The adjusted measurement is obtained by removing the expected value, i.e.,
$\text{Adj}\AveP = \frac{\AveP-E[\AveP]}{1-E[\AveP]}=\frac{\AveP-|O|/N}{1-|O|/N}$,
such that a value of 0 is the expected score of a random result and 1 is optimal.
Average Precision is closely related to the area under the precision recall curve (\AUPRC) measure,
but is more efficient to compute and does not have the definition problems of the latter.
\AUROC{}, the area under the receiver operating curve is the standard measure in evaluating outlier detection
methods, corresponding to the probability of ranking an outlier $o$ before an inlier $i$,
i.e., $\AUROC{=}P(\rank(o){<}\rank(i))$.
But this measure has been criticized for implying a uniform weighting over all ranks.
Readers from machine learning will likely be more comfortable with \AveP{},
while data mining researchers tend to be more used to \AUROC{}.
A discussion of evaluation metrics can be found in Campos et al.~\cite{DBLP:journals/datamine/CamposZSCMSAH16}.

We first give an overview of the experiments:
First, the bandwidth heuristics to choose $\gamma$ are compared, followed by a perturbation experiment of this parameter.
Afterward, %
we compare the proposed LO-SVM to the baselines of OCSVM and SVDD, and additionally the KNN with $K{=}1$ baseline.
Last, we evaluate the effect of batchwise removal compared to removing one at a time. 

\begin{figure}[tb]
  \begin{subfigure}[b]{0.5\linewidth}
  \centering
  \includegraphics[width=\linewidth]{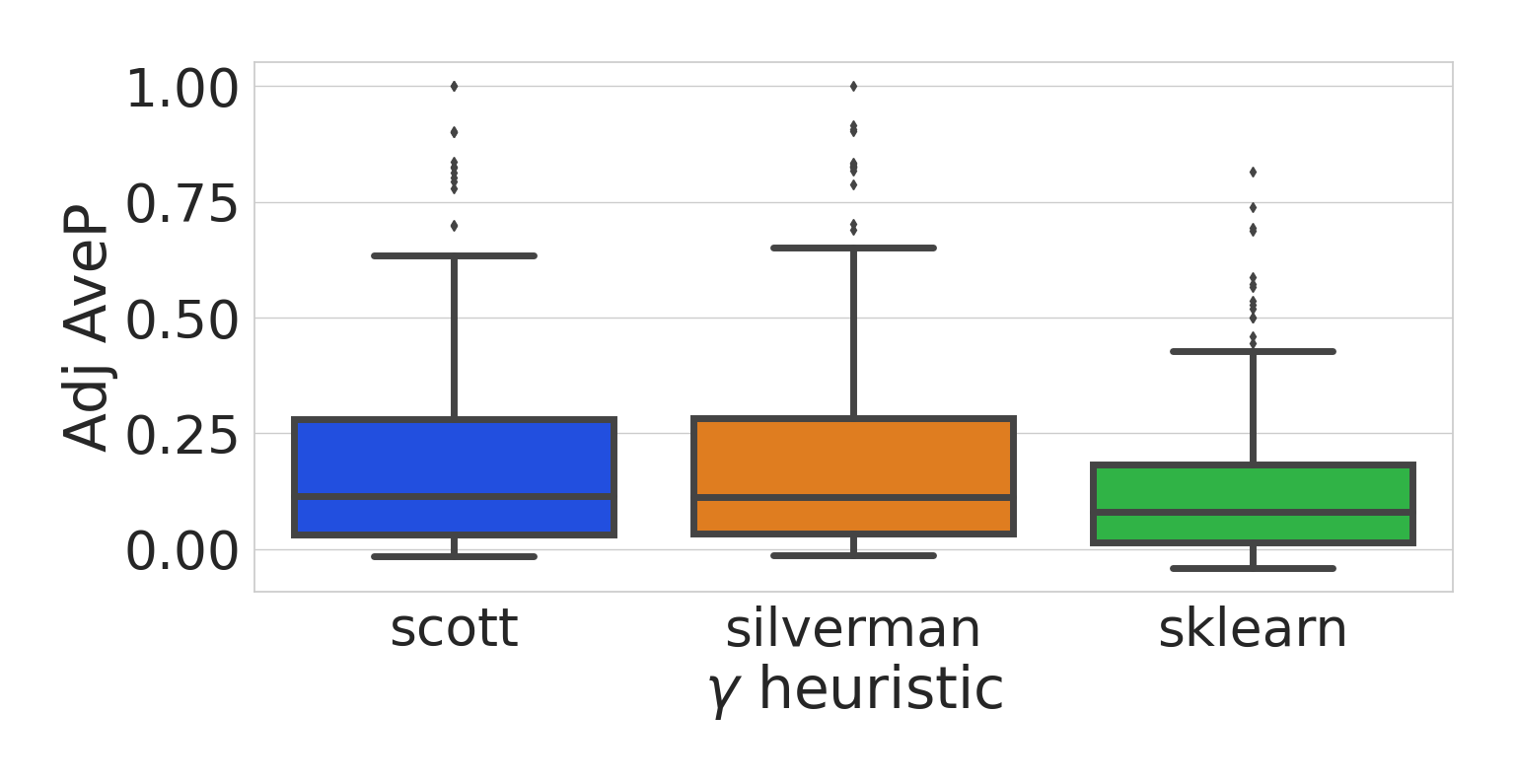}
  \end{subfigure}%
  \hfill
  \begin{subfigure}[b]{0.5\linewidth}
  \centering
  \includegraphics[width=\linewidth]{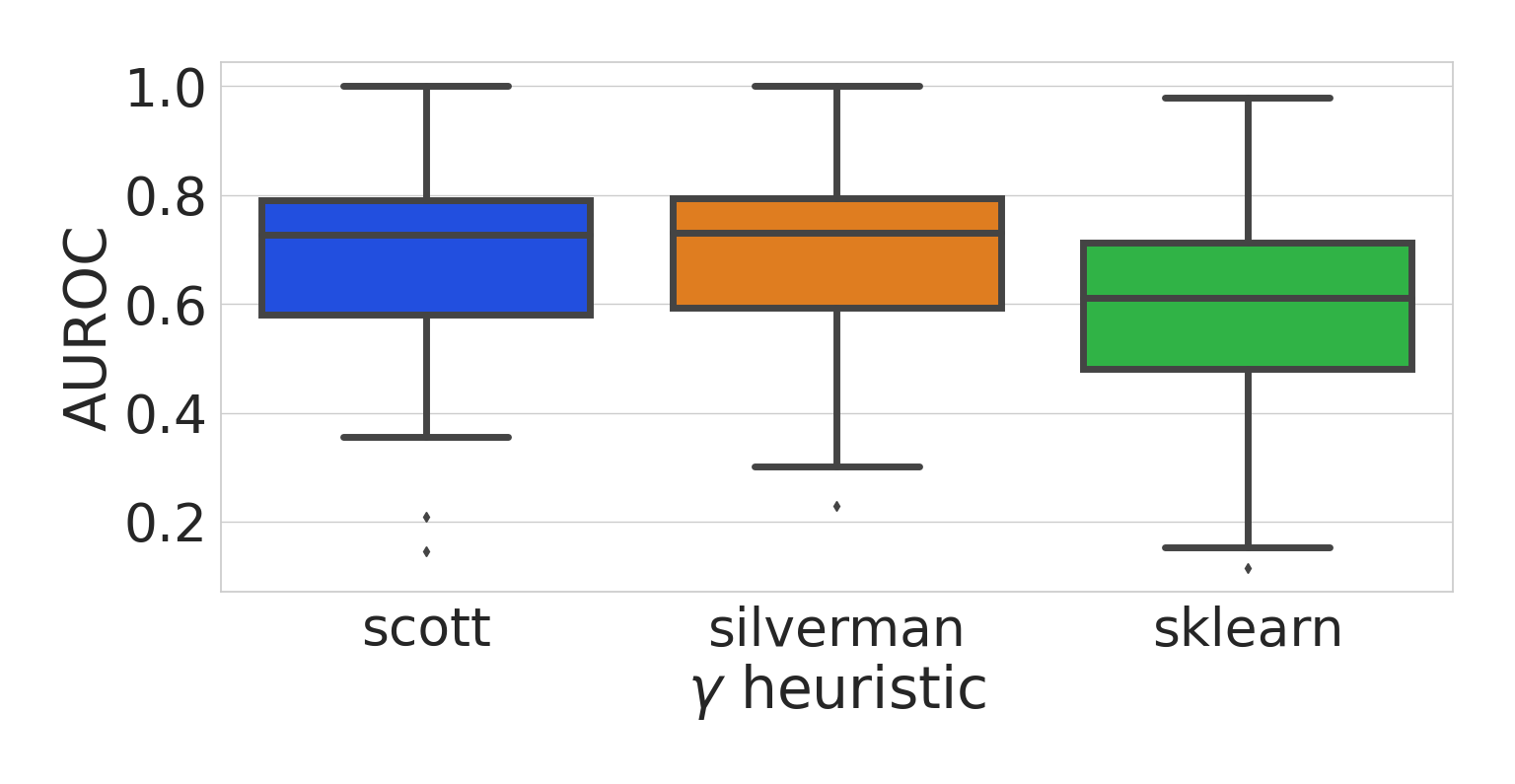}
  \end{subfigure}
  \caption{Comparison of bandwidth heuristics averaged over all data sets with all outlier percentage and all versions,
  using LOSDD, $C=1$ and no batching.}
  \label{fig:bandwidth}
\end{figure}

\reffig{fig:bandwidth} compares the three popular unsupervised heuristics for choosing the bandwidth $\gamma$ parameter.
Scott's and Silverman's rule of thumb perform very similar, while the sklearn heuristic that does not take the data set size into
account appears to be too simple and performs much worse.
Because of the slightly better median performance of Silverman's rule of thumb, we will in the following use this to choose $\gamma$.

\begin{figure}[tb]
  \begin{subfigure}[b]{0.5\linewidth}
  \centering
  \includegraphics[width=\linewidth]{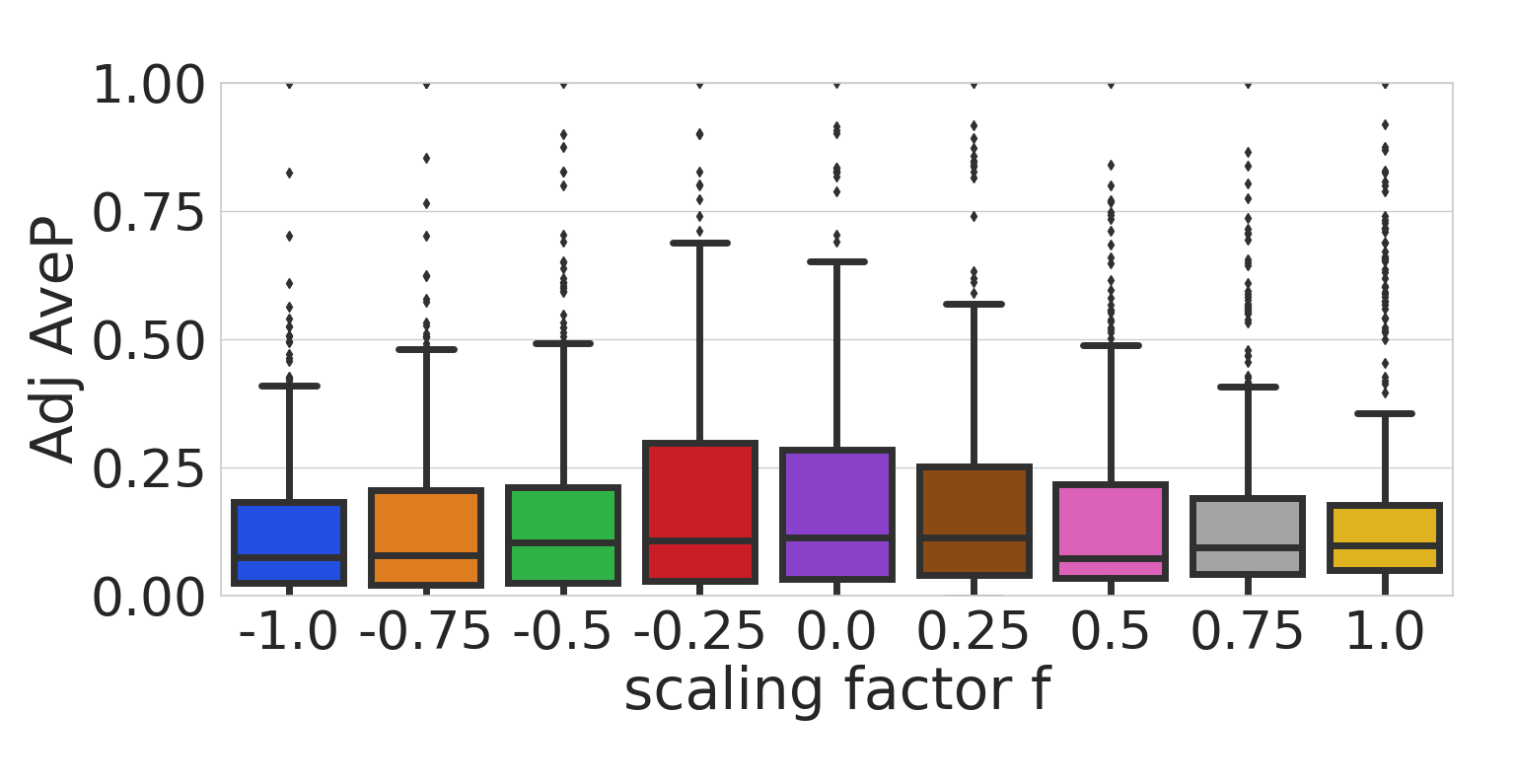}
  \end{subfigure}%
  \hfill
  \begin{subfigure}[b]{0.5\linewidth}
  \centering
  \includegraphics[width=\linewidth]{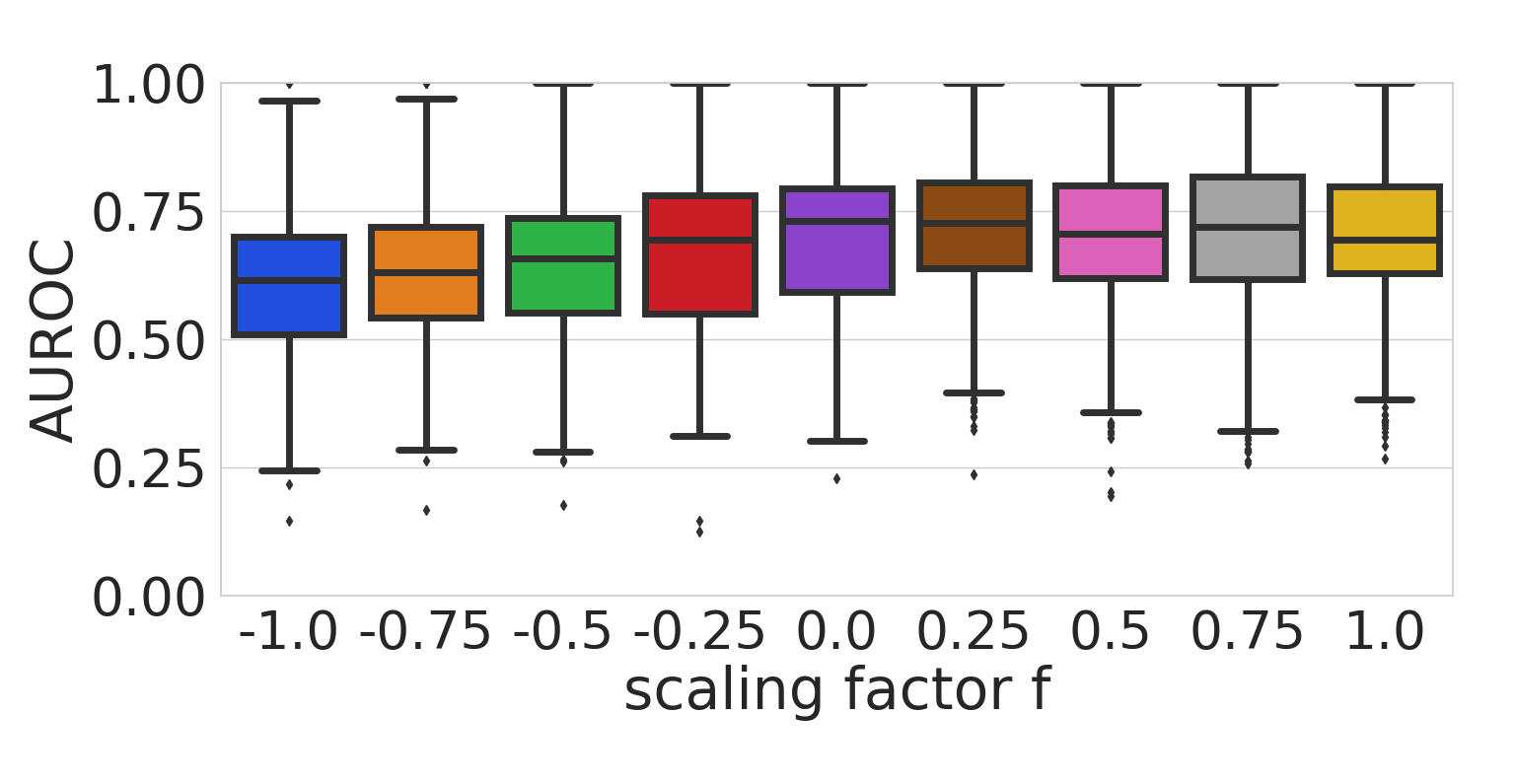}
  \end{subfigure}
  \caption{AdjAveP and \AUROC{} averaged over all data sets with LOSDD and $\gamma = 10^f \cdot \gamma_s$, with the standard Silverman at $f=0$.}
  \label{fig:scalingfactor}
\end{figure}

To further verify the choice of $\gamma$ with this heuristic, we experiment with varying the $\gamma$ parameter
using $\gamma {=} 10^f {\cdot} \gamma_s$ and $f$ from -1 to 1 in steps of 0.25,
and $\gamma_s$ is the base value obtained by Silverman's rule.
As seen in \reffig{fig:scalingfactor}, the value chosen by Silverman's rule is reasonable as a default value.

\begin{figure*}[tb]
  \begin{subfigure}[b]{\linewidth}
  \centering
  \includegraphics[trim={0.39cm 6.5cm 0.49cm 7.45cm},clip,width=\textwidth]{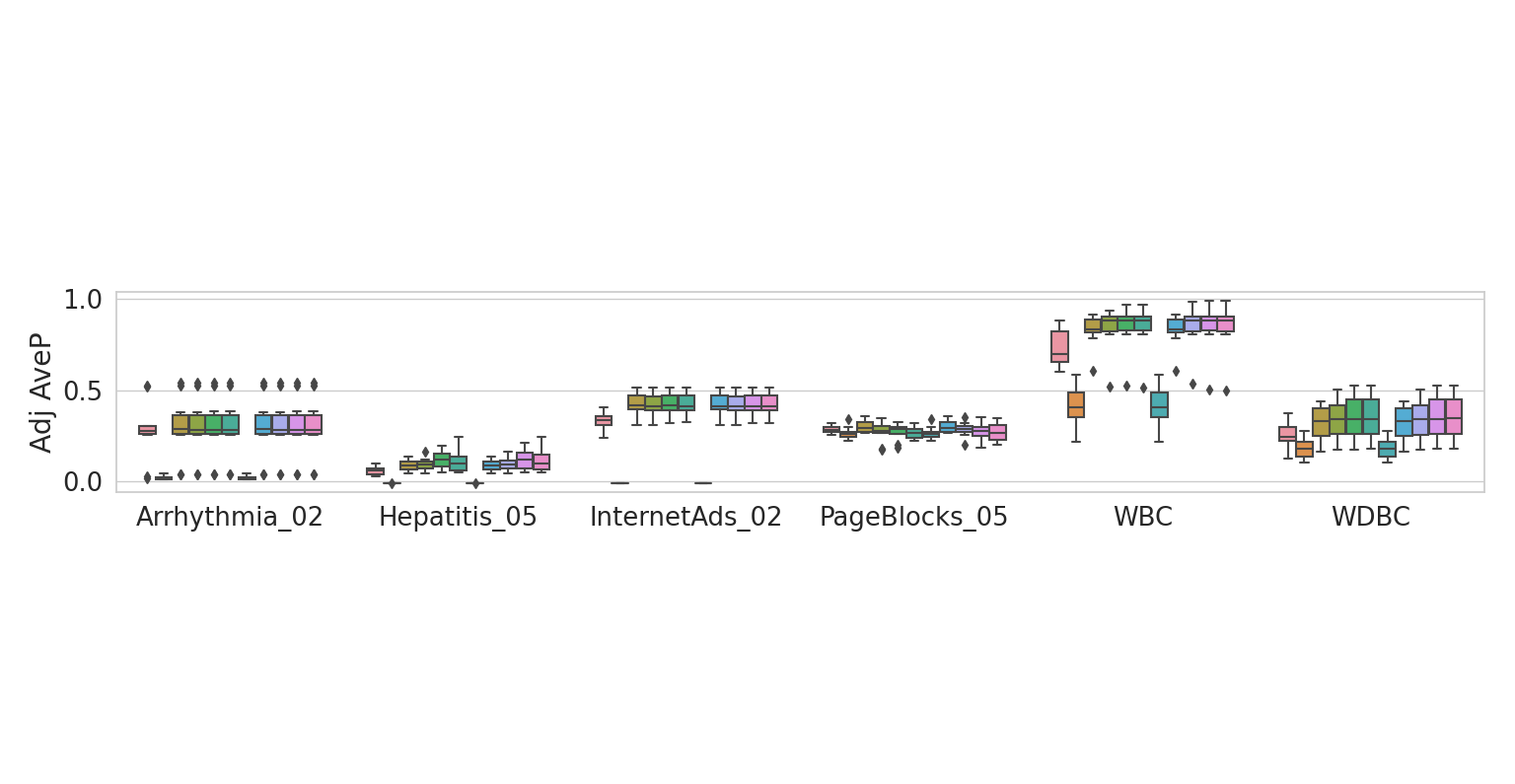}
  \end{subfigure}
  \begin{subfigure}[b]{\linewidth}
  \centering
  \includegraphics[trim={0.39cm 6.6cm 0.49cm 7.2cm},clip,width=\textwidth]{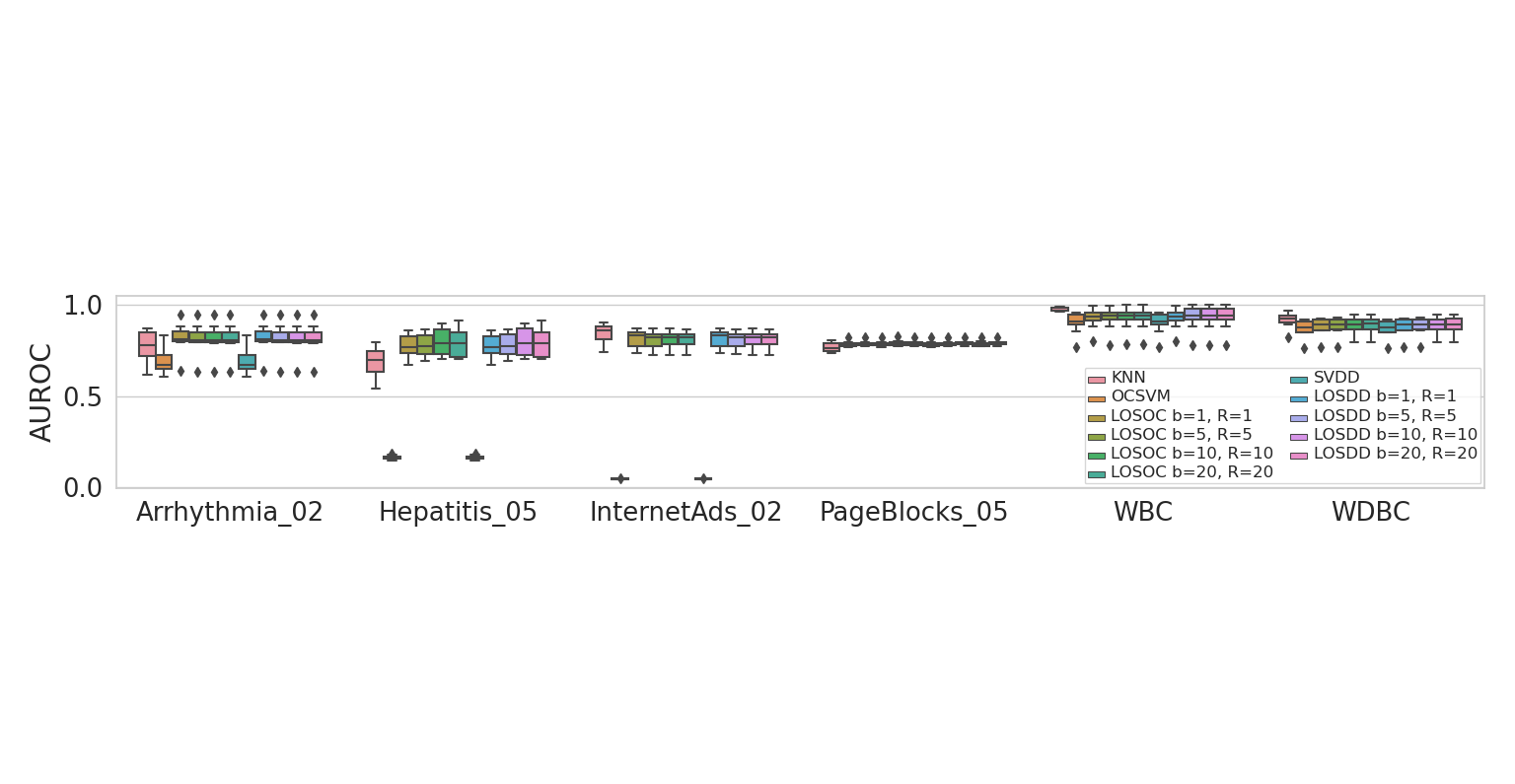}
  \end{subfigure}
  \caption{Increase of $b,R$ under \AdjAveP{} and \AUROC{} averaged over all versions. Silverman's r.o.t., $C=1$.}
  \label{fig:numbatch}
\end{figure*}

In \reffig{fig:numbatch}, we now compare the LO-SVM approach over the baseline
methods such as nearest neighbor outlier detection, OCSVM, and SVDD.
We observe that most of the time, LO-SVM performs better than the baseline KNN method
with respect to average precision (but not always with respect to AUROC),
whereas the original OCSVM and SVDD approaches perform worst (and even struggle badly
with some of the data sets), indicating the presence of masking in these data sets.
The batchwise removal of 1, 5, 10 and 20 outliers at once does not appear to have a negative
effect on AUROC, except on the PageBlocks data set; and only a slight negative effect on the average precision.
The mean results are also given in \reftab{tab:intexttable-adjusted-average-precision}.
In this presentation, the difference between the rankings obtained by average precision (the adjustment does not change the rankings)
and AUROC becomes most apparent: while KNN performs best on 3 of these data sets with respect to AUROC,
the average precision scores of the LO-SVM approaches are clearly better.

\begin{figure}[!tb]
  \begin{subfigure}[b]{\linewidth}
  \centering
  \includegraphics[width=\linewidth]{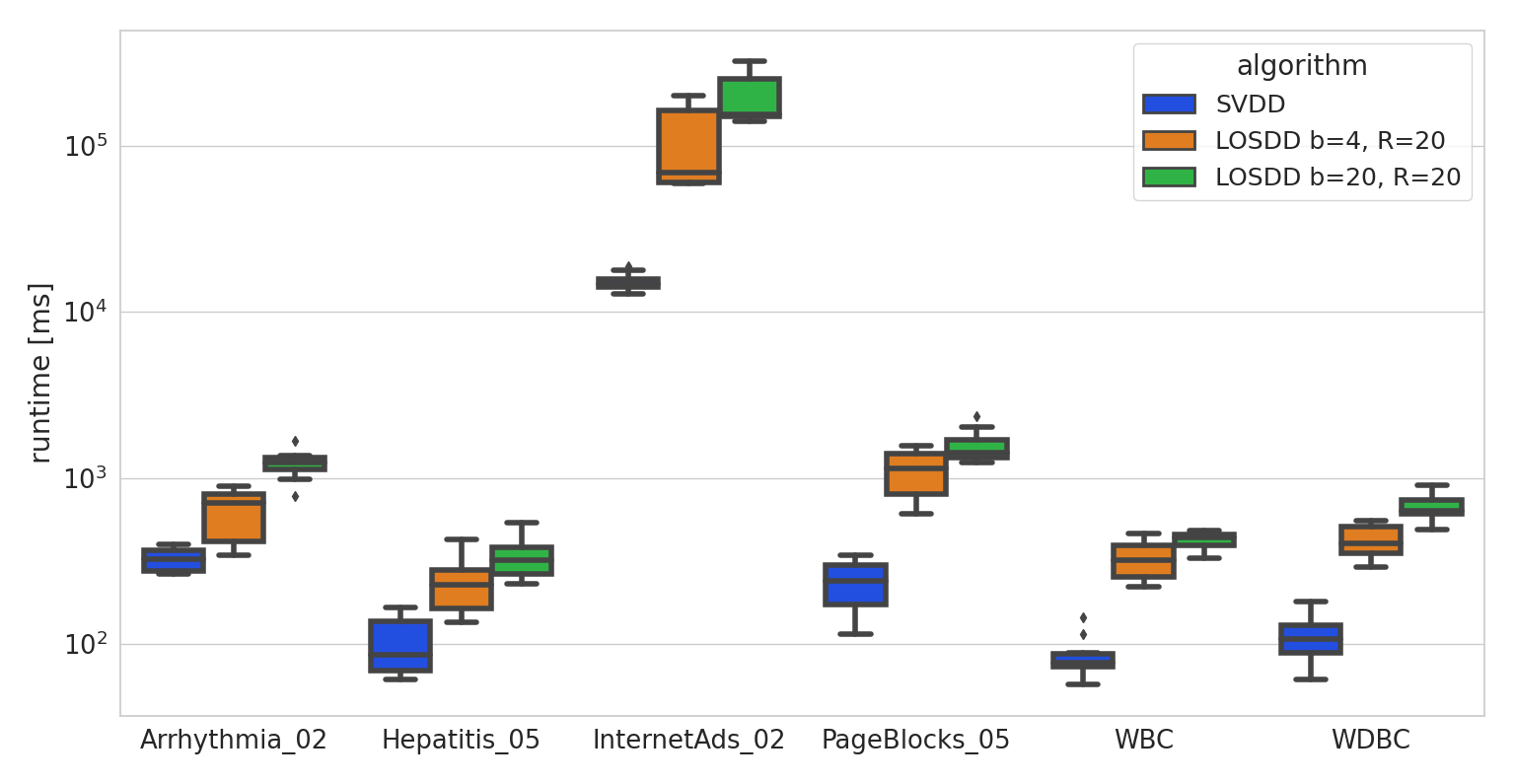}
  \end{subfigure}
  \begin{subfigure}[b]{\linewidth}
  \centering
  \includegraphics[width=\linewidth]{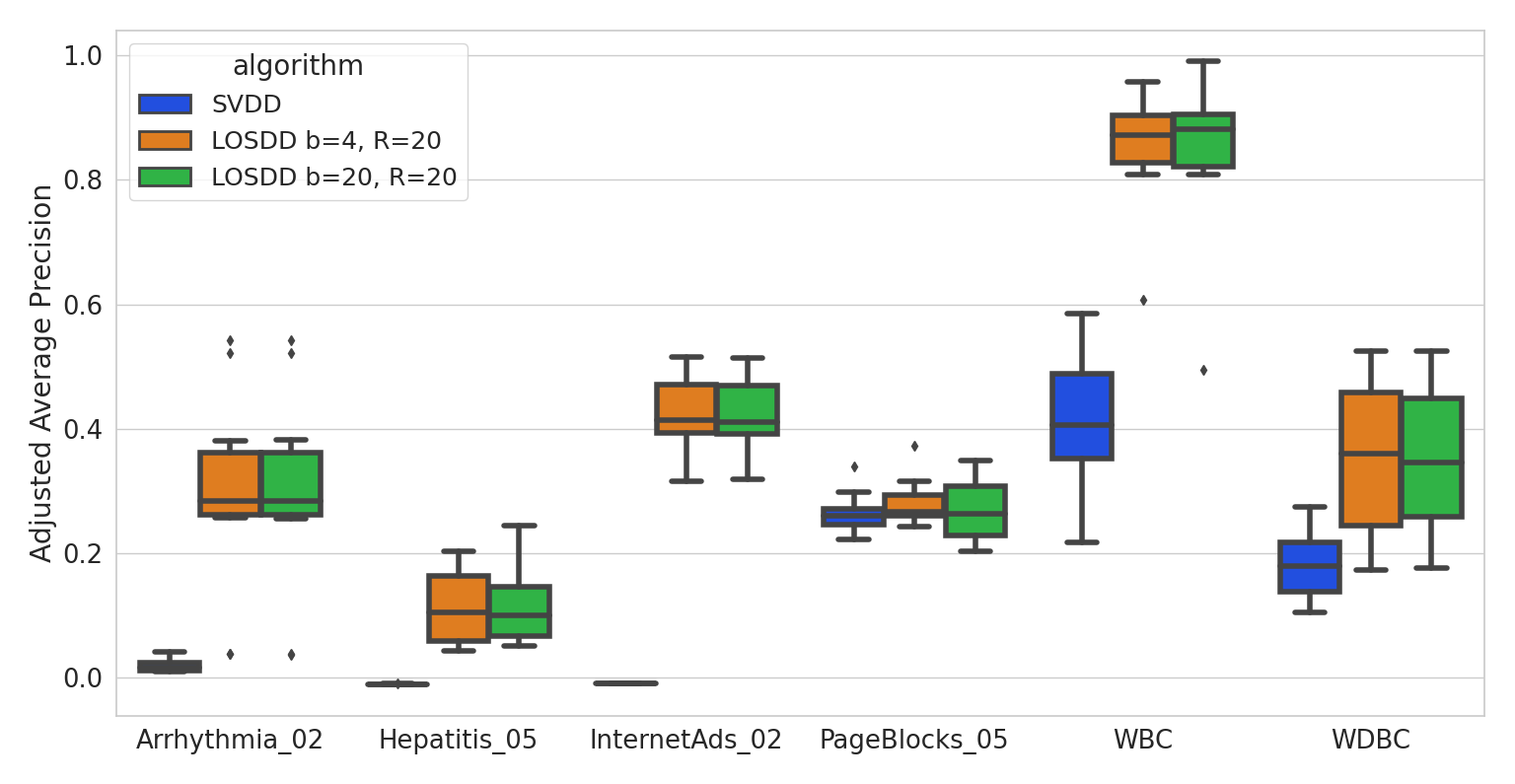}
  \end{subfigure}
  \begin{subfigure}[b]{\linewidth}
  \centering
  \includegraphics[width=\linewidth]{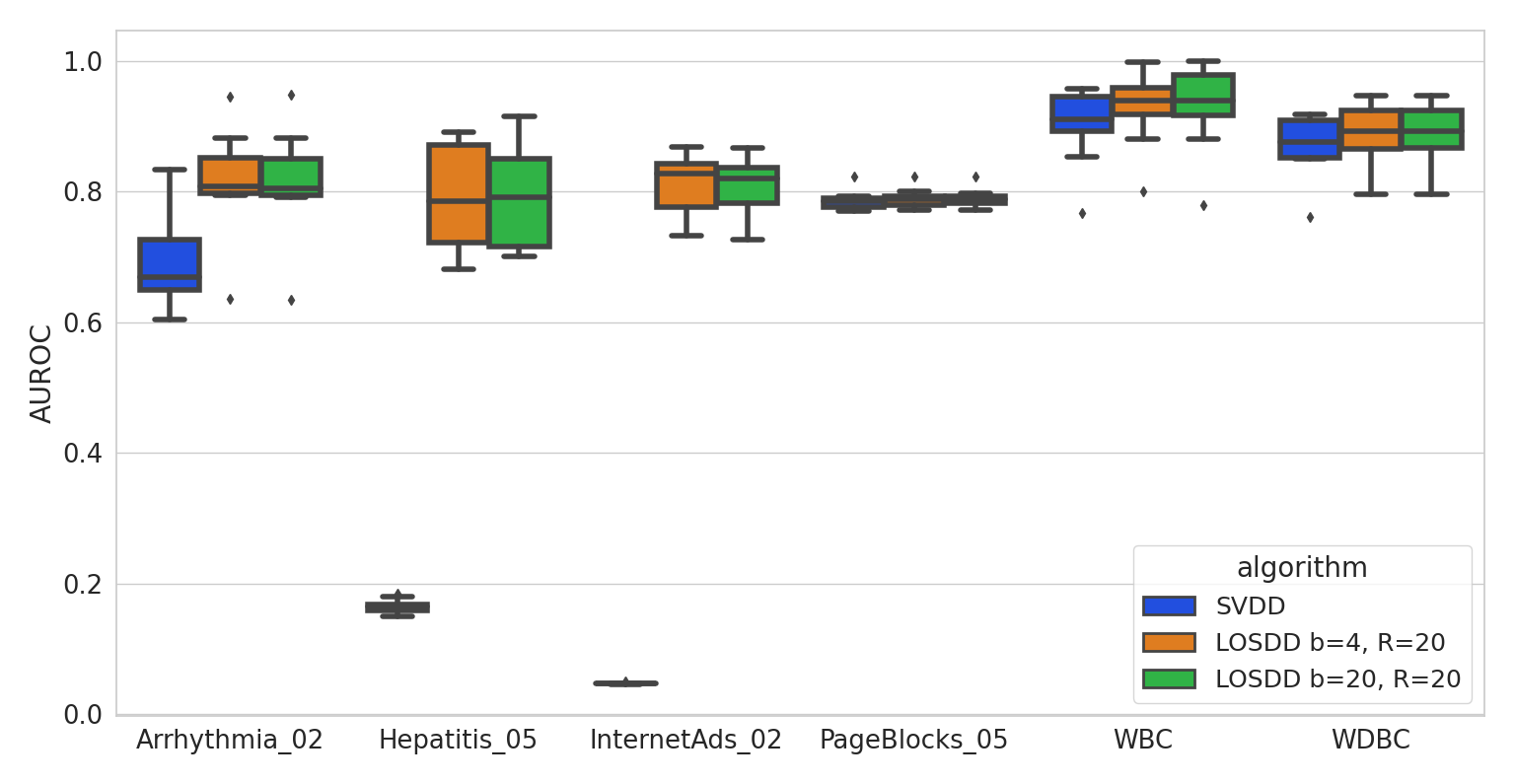}
  \end{subfigure}
  \caption{Runtime and performance when removing $R=20$ SVs with $b=4$, and $R=20$ SVs with $b=20$.}
  \todo[inline]{Gehört hier nicht überall konsequenterweise eine 0 times 0 Variante dazu, wo nur LOO-Evaluation stattfindet?}
  \label{fig:runtimebatchsize}
\end{figure}

In the last experiment, we evaluate the benefits of batchwise removal.
We remove 20 support vectors each, but once in 4 batches of 5 SVs each, and then 20 times a single SV, retraining inbetween
of each batch.
As seen in \reffig{fig:runtimebatchsize}, the runtime improves with larger batch sizes at no noticeable loss in quality.

\begin{table*}[tb]
\caption{Mean Adj AveP \& Mean AUROC averaged over versions. Silverman's rule of thumb, $C=1$ are used.}
\label{tab:intexttable-adjusted-average-precision}
\centering
\begin{tabular}{lllllll}
Mean Adj AveP &   \rotatelow{Arrhythmia\_02}\hspace*{-1cm} &    \rotatelow{Hepatitis\_05}\hspace*{-1cm} &  \rotatelow{InternetAds\_02}\hspace*{-1cm} &   \rotatelow{PageBlocks\_05}\hspace*{-1cm} &             \rotatelow{WBC}\hspace*{-1cm} &            \rotatelow{WDBC} \\
\hline
KNN            &           0.278 &           0.059 &           0.329 &           0.285 &           0.733 &           0.259 \\
OCSVM          &           0.019 &           -0.01 &          -0.009 &           0.265 &           0.417 &           0.178 \\
LOSOC $b=1, R=1$       &  \textbf{0.293} &           0.087 &           0.423 &  \textbf{0.300} &           0.826 &           0.318 \\
LOSOC $b=5, R=5$       &  \textbf{0.293} &           0.094 &           0.421 &           0.272 &           0.838 &           0.337 \\
LOSOC $b=10, R=10$      &  \textbf{0.293} &           0.117 &           0.422 &            0.27 &           0.844 &           0.347 \\
LOSOC $b=20, R=20$      &           0.292 &           0.111 &  \textbf{0.424} &           0.267 &           0.843 &  \textbf{0.349} \\
SVDD           &           0.019 &           -0.01 &          -0.009 &           0.266 &           0.417 &           0.178 \\
LOSDD $b=1, R=1$  &  \textbf{0.293} &           0.087 &           0.423 &  \textbf{0.300} &           0.826 &           0.318 \\
LOSDD $b=5, R=5$  &  \textbf{0.293} &           0.095 &           0.421 &           0.285 &  \textbf{0.848} &           0.336 \\
LOSDD $b=10, R=10$ &  \textbf{0.293} &  \textbf{0.123} &           0.422 &           0.273 &  \textbf{0.848} &           0.348 \\
LOSDD $b=20, R=20$ &           0.292 &           0.114 &           0.422 &           0.268 &           0.845 &  \textbf{0.349} \\
\hline
\hline
Mean AUROC &    &  &  &    &   &               \\
\hline
KNN            &           0.772 &           0.688 &  \textbf{0.845} &           0.768 &  \textbf{0.975} &  \textbf{0.918} \\
OCSVM          &            0.69 &           0.165 &           0.047 &           0.786 &           0.903 &           0.872 \\
LOSOC $b=1, R=1$       &  \textbf{0.817} &           0.769 &           0.816 &           0.788 &           0.927 &           0.883 \\
LOSOC $b=5, R=5$       &           0.815 &           0.781 &            0.81 &           0.788 &           0.928 &           0.885 \\
LOSOC $b=10, R=10$      &           0.815 &           0.797 &            0.81 &           0.789 &           0.929 &           0.889 \\
LOSOC $b=20, R=20$      &           0.814 &           0.791 &            0.81 &           0.789 &           0.929 &            0.89 \\
SVDD           &            0.69 &           0.165 &           0.047 &           0.786 &           0.903 &           0.872 \\
LOSDD $b=1, R=1$  &  \textbf{0.817} &           0.769 &           0.816 &           0.788 &           0.927 &           0.882 \\
LOSDD $b=5, R=5$ &           0.815 &           0.782 &            0.81 &           0.789 &           0.932 &           0.885 \\
LOSDD $b=10, R=10$ &           0.815 &  \textbf{0.799} &            0.81 &           0.788 &           0.932 &           0.889 \\
LOSDD $b=20, R=20$ &           0.814 &           0.792 &           0.809 &  \textbf{0.790} &           0.932 &            0.89 \\
\end{tabular}
\end{table*}

\section{Conclusion}
We observed that the one-class support vector machines are designed assuming ``pure'' data sets,
and perform poorly on ``dirty'' input data, because the optimum model is strongly influenced by outliers.
Because of this, outliers are likely to become support vectors, and influence the decision surface of the model a lot.
To alleviate this, we developed LO-SVM, a leave-one-out SVM for the dirty data scenario that evaluates
each object with respect to the remainder of the data set.
Furthermore, we can prune the model by removing the top detected outliers, and only fit the majority of the data.
Naively, this would require training $N$ support vector machines, which makes the naive approach infeasible.
As one outlier may mask another, we may also want to repeat the removal procedure multiple times.
To solve this, we show how to accelerate this using a short retraining of the existing SVM,
and that we only need to consider the current support vectors.
For further acceleration, we investigate batchwise removal.

\setlength\bibitemsep{0pt}
\AtNextBibliography{\small}
\printbibliography %

\end{document}


\title{Supplementary Material LOSEVM}
\subtitle{\Large LOSDD: Leave-Out Support Vector Data Description for Outlier Detection}
\author{Daniel Boiar\orcidID{0000-0002-6856-9848} \and Thomas Liebig\orcidID{0000-0002-9841-1101} \and Erich Schubert\orcidID{0000-0001-9143-4880}}
\institute{TU Dortmund University, Dortmund, Germany\\
\email{\{daniel.boiar,thomas.liebig,erich.schubert\}@tu-dortmund.de}}

\maketitle

\section{Proof Theorem 1}
\begin{proof}
 The object function is optimal for $\alpha$ under the constraint and is given by
 \begin{align*}
 OPT = \sum\limits_{i=1}^N\sum\limits_{j=1}^N \alpha_i\alpha_jK(x_i,x_j)-\sum\limits_{i=1}^N\alpha_iK(x_i,x_i)
 \; ,
 \end{align*}
 whereas being subject to 
 \begin{align*}
   0\leq \alpha_i\leq C \ \forall i\in\{1,\dots,N\},\sum\nolimits_{i=1}^N \alpha_i = 1.
 \end{align*}
 In particular, only $\alpha_i\neq 0$ is accumulated. 
 Since it was assumed that the left out point satisfies $\alpha_t=0$, it follows $OPT = OPT_t$, where 
 \begin{align*}
   OPT_t = \sum\limits_{\substack{i=1 \\ i\neq t}}^N\sum\limits_{\substack{j=1 \\ j\neq t}}^N \alpha_i\alpha_jK(x_i,x_j)-\sum\limits_{\substack{i=1 \\ i\neq t}}^N\alpha_iK(x_i,x_i)
   \;.
 \end{align*}
 Its constraints are also met, since $\alpha_t=0$. 
 The value $OPT_t$ cannot be improved by changing $\alpha$, otherwise, $OPT$ could be improved as well. 
 This would contradict the optimality, so $OPT_t$ is optimal.
\end{proof}